\documentclass[journal]{IEEEtran}

\usepackage[T1]{fontenc}
\usepackage[utf8]{inputenc}

\usepackage{amsmath}

\usepackage{graphicx}
\usepackage{multirow}
\usepackage[flushleft]{threeparttable}

\usepackage{cite}
\PassOptionsToPackage{hidelinks}{hyperref}
\usepackage{hyperref}

\usepackage{siunitx}
\DeclareSIUnit{\tokens}{tokens}
    
\begin{document}

\title{ProfileXAI: User‑Adaptive Explainable AI}

\newcommand{\authcell}[2]{%
  \begin{tabular}{@{}p{.47\textwidth}@{}}
  \centering \textbf{#1}\\[0.2ex]
  #2
  \end{tabular}
}

\author{%
  Gilber~A.~Corrales,
  Carlos~Andr\'es~Ferro~S\'anchez,
  Reinel~Tabares\mbox{-}Soto,
  Jes\'us~Alfonso~L\'opez~Sotelo,
  Gonzalo~A.~Ruz,
  and~Johan~Sebastian~Pi\~na~Dur\'an%

    \thanks{Manuscript received July~27,~2025; accepted August~13,~2025.
  This work was supported in part by the graduate assistantship scholarship granted through
  Resolution No.~7906 by the Vice-Rectorate for Research and the Universidad Autónoma de Occidente;
  the National Agency for Research and Development (ANID), Applied Research Subdirection/IDeA
  I\,+\,D 2023 Grant [folio ID23110357]; \emph{Classification of Alzheimer's stages using Nuclear
  Magnetic Resonance Imaging and clinical data from Deep Learning techniques} [873--139],
  Universidad Autónoma de Manizales, Manizales, Colombia; ORIGEN~0011323, Sistema General de Regalías (SGR)---Asignación para la Ciencia, Tecnología e Innovación project BPIN~2021000100368;
  \emph{Technological platform for the classification of Alzheimer's disease stages using nuclear magnetic resonance imaging, clinical data and Deep Learning techniques} PRY-89, Universidad de Caldas, Manizales, Colombia; \emph{Estrategia didáctica interactiva virtual para el fomento de habilidades TIC y su relación con el pensamiento computacional} PRY-121, Universidad de Caldas, Manizales, Colombia.
  G.~A.~R. thanks ANID FONDECYT~1230315, ANID-MILENIO-NCN2024\_103, ANID PIA/BASAL~AFB240003, and
  Centro de Modelamiento Matemático (CMM)~FB210005, BASAL funds for centers of excellence from ANID--Chile.
  (Corresponding author: Gilber A.~Corrales.)}

  \thanks{G.~A.~Corrales is with the Facultad de Ingeniería y Ciencias Básicas, Universidad Autónoma de Occidente, Cali 760000, Colombia; and with the Escuela de Gobierno, GobLab, Universidad Adolfo Ibáñez, Santiago de Chile 8320000, Chile. (e-mail: gacorrales@uao.edu.co).}%
  \thanks{C.~A.~Ferro~S\'anchez is with the Facultad de Ingeniería y Ciencias Básicas, Universidad Autónoma de Occidente, Cali 760000, Colombia.}%
  \thanks{R.~Tabares\mbox{-}Soto is with the Escuela de Gobierno, GobLab, Universidad Adolfo Ibáñez, and the Facultad de Ingeniería y Ciencias, Universidad Adolfo Ibáñez, Santiago de Chile 8320000, Chile; and with the Departamento de Electrónica y Automatización, Universidad Autónoma de Manizales, and the Departamento de Sistemas e Informática, Universidad de Caldas, Manizales 170003, Colombia.}%
  \thanks{J.~A.~López~Sotelo is with the Facultad de Ingeniería y Ciencias Básicas, Universidad Autónoma de Occidente, Cali 760000, Colombia.}%
  \thanks{G.~A.~Ruz is with the Facultad de Ingeniería y Ciencias, Universidad Adolfo Ibáñez; Millennium Nucleus for Social Data Science (SODAS); and Center of Applied Ecology and Sustainability (CAPES), Santiago de Chile 8320000, Chile.}%
  \thanks{J.~S.~Pi\~na~Dur\'an is with the Escuela de Gobierno, GobLab, Universidad Adolfo Ibáñez, Santiago de Chile 8320000, Chile; and with Universidad Autónoma de Manizales, Manizales 170003, Colombia.}%
}

\maketitle

\begin{abstract}
ProfileXAI is a model- and domain-agnostic framework that couples post-hoc explainers (SHAP, LIME, Anchor) with retrieval - augmented LLMs to produce explanations for different types of users. The system indexes a multimodal knowledge base, selects an explainer per instance via quantitative criteria, and generates grounded narratives with chat-enabled prompting. On Heart Disease and Thyroid Cancer datasets, we evaluate fidelity, robustness, parsimony, token use, and perceived quality. No explainer dominates: LIME achieves the best fidelity--robustness trade-off (Infidelity $\le 0.30$, $L<0.7$ on Heart Disease); Anchor yields the sparsest, low-token rules; SHAP attains the highest satisfaction ($\bar{x}=4.1$). Profile conditioning stabilizes tokens ($\sigma \le 13\%$) and maintains positive ratings across profiles ($\bar{x}\ge 3.7$, with domain experts at $3.77$), enabling efficient and trustworthy explanations.
\end{abstract}

\section{Introduction}

Artificial intelligence (AI) permeates most processes \cite{ASHOK2022102433,cath2018governing}. As model architectures and parameter counts soar, their decision mechanisms become opaque, effectively turning them into black‑box systems \cite{Manure2023,10.1007/978-3-031-59216-4_9}. Explainable AI (XAI) aims to restore transparency while maintaining predictive accuracy \cite{minh2022explainable}; however, existing techniques often fail to adapt their explanations to audiences with heterogeneous expertise \cite{frasca2024explainable,valiente2023shap}. Integrating classical XAI with large language models (LLMs) has recently emerged as a promising \cite{mavrepis2024xaialllargelanguage}.

Initial case studies utilized ChatGPT to generate SHAP and counterfactual outputs for student-risk analytics and the Iris benchmark \cite{susnjak2024beyond}. Spitzer et al. later demonstrated that context-augmented prompting yields higher user satisfaction than retrieval-based prompting when explaining a deep-learning cost predictor \cite{10.1145/3670653.3677488}. In the networking domain, a fully automated 6G framework integrates XGBoost-SHAP with Llama 2 to diagnose SLA-latency anomalies, thereby increasing operator trust while revealing occasional decision errors \cite{10742571}. Complementary methodological work formalises evaluation metrics—soundness, completeness, and fluency—and demonstrates that human readers prefer narrative SHAP summaries \cite{zytek2024llmsxaifuturedirections}. A recent survey synthesises these advances but highlights persistent issues of coherence and factuality \cite{bilal2025llmsexplainableaicomprehensive}. Operational prototypes illustrate the practical upside: TalkToModel embeds GPT-J/3.5 within an interactive dialogue engine that reformats attribution-based explanations on demand, and most clinicians and ML professionals prefer it to conventional dashboards \cite{slack2023talktomodelexplainingmachinelearning}. In recommender systems, LLM-generated justifications significantly enhance perceived transparency across feature, collaborative, and knowledge-based pipelines \cite{10.1145/3631700.3665185}. Finally, explanation-consistency finetuning improves the logical alignment of LLM summaries by approximately 10 \% without degrading task accuracy \cite{chen2024consistentnaturallanguageexplanationsexplanationconsistency}.

We present a domain- and model-agnostic framework that advances this literature along three key axes. First, it adapts output granularity and style to distinct user profiles — machine learning experts, domain experts, and non-technical users — thereby maximizing relevance. Second, an interactive chat module enables stakeholders to refine queries and resolve residual uncertainties in real-time. Third, a dynamic engine selects the most suitable XAI method and enriches it through a multimodal retrieval-augmented generation pipeline, producing grounded, audience-specific narratives.

\section{Methodology}

\subsection{Pipeline}

\begin{figure*}[!t]
\centering
\includegraphics[width=\textwidth]{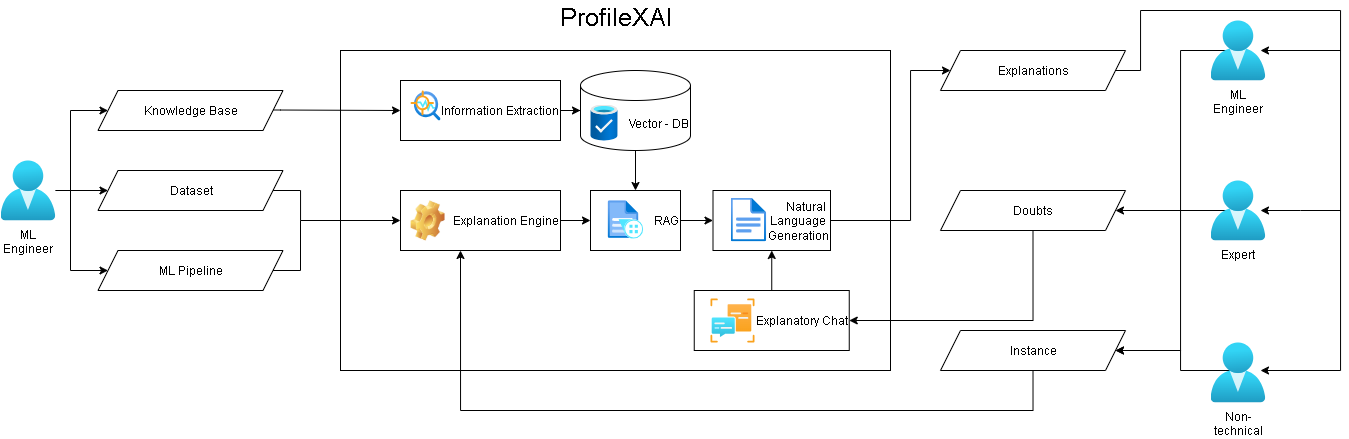}
\caption{ProfileXAI system architecture.}
\label{fig:ProfileXAI}
\end{figure*}

Figure~\ref{fig:ProfileXAI} illustrates the ProfileXAI architecture. The user—a machine‑learning (ML) engineer—provides three inputs:

\begin{itemize}
\item \textbf{Knowledge base}, which supplies contextual information used to enrich the explanations, and this can be multimodal.

\item \textbf{Black‑box model} whose behaviour is to be explained (ex., Support vector machine, Multilayer perceptron, Random forest).

\item \textbf{Dataset} (or a subset thereof) on which the model operates.
\end{itemize}

The information‑extraction module then processes the knowledge base in a multimodal manner, identifies the most relevant components (extracting images or text from different types of documents), and stores them in a vector database. When an instance is submitted, the Retrieval-Augmented Generation (RAG) subsystem retrieves relevant fragments from the database to compose the generation prompt, enabling the system to generate explanations in natural language with context. The Explanation Engine, based on the instance entered by the user, executes three interpretability methods: SHAP \cite{NIPS2017_8a20a862}, LIME \cite{ribeiro2016whyitrustyou}, and Anchor \cite{Ribeiro_Singh_Guestrin_2018}, and automatically selects the most suitable one for each instance according to predefined metrics based on some metrics of \cite{PAWLICKI2024128282}.

The resulting explanation is produced in natural language and tailored to three user profiles:
\begin{itemize}
\item \textbf{ML engineer}: technical details, performance metrics, and raw model and explanation outputs.
\item \textbf{Domain expert}: a translation of the explanatory content into terminology aligned with the application domain.
\item \textbf{Non‑technical user}: accessible language with illustrative examples and minimal jargon.
\end{itemize}

If the user poses follow-up questions, the interactive chat module enables a deeper exploration of any aspect of the generated explanation.

\subsection{Experiments}

We conducted experiments on two public datasets: Heart Disease with 13 features \cite{heart_disease_45} and Differentiated Thyroid Cancer Recurrence with 16 features \cite{differentiated_thyroid_cancer_recurrence_915}. The knowledge base comprised the articles \cite{Detrano1989InternationalAO,Borzooei2024Thyroid} . We trained a multilayer perceptron (MLP) on the first dataset and a Random Forest on the second. Our evaluation comprised three blocks:

\begin{enumerate}
    \item \textbf{XAI‑metric analysis.}
    We adapted and assessed three standard interpretability metrics —Infidelity \cite{10.5555/3454287.3455271}, Lipschitz \cite{10.5555/3327757.3327875}, and Effective complexity \cite{nguyen2020quantitativeaspectsmodelinterpretability} —across 100 instances of each dataset. For every explanation method, we report the mean and standard deviation of each metric Table~\ref{tab:xai_metrics}.
    
    \item \textbf{Token consumption.}  
    We recorded the number of tokens consumed per user profile (ML engineer, domain expert, non‑technical) and per explanation method on 200 instances of each dataset. Table \ref{tab:tokens} summarises the averages.
    
    \item \textbf{Satisfaction simulation.}  
    Following the Hoffman survey \cite{10.3389/fcomp.2023.1096257}, a simulated LLM scored seven explanation-quality items on a 1–5 scale (1 = very low, 5 = very high). We assessed 200 instances per Dataset, stratifying the results by user profile and explanation method. We thus obtained an average satisfaction score for each profile Table~\ref{tab:quality_metrics}.

\end{enumerate}

\section{Results and analysis}
\begin{table*}[!t]
\caption{Mean $\pm$ standard deviation of three XAI metrics—Infidelity, Lipschitz, and Effective Complexity (EffComp)—computed over 100 instances for each explanation method (Anchor, LIME, SHAP) on the Heart Disease dataset (Dataset~A) and the Differentiated Thyroid Cancer Recurrence dataset (Dataset~B). Cells marked “--” indicate that the metric is not well-defined for Anchor.}
\label{tab:xai_metrics}
\centering
\renewcommand{\arraystretch}{1.15}
\setlength{\tabcolsep}{4pt}
\begin{threeparttable}
\begin{tabular}{|l|*{3}{r@{\,±\,}l|}*{3}{r@{\,±\,}l|}}
\hline
\multirow{2}{*}{\textbf{Method}}
  & \multicolumn{6}{c|}{\textbf{Dataset A}}
  & \multicolumn{6}{c|}{\textbf{Dataset B}} \\
\cline{2-13}
  & \multicolumn{2}{c|}{\textbf{Infidelity}}
  & \multicolumn{2}{c|}{\textbf{Lipschitz}}
  & \multicolumn{2}{c|}{\textbf{EffComp}}
  & \multicolumn{2}{c|}{\textbf{Infidelity}}
  & \multicolumn{2}{c|}{\textbf{Lipschitz}}
  & \multicolumn{2}{c|}{\textbf{EffComp}} \\
\hline
Anchor 
  & -- & --
  & 0.88 & 0.29
  & 3.48 & 4.69
  & -- & --
  & 1.49 & 0.66
  & 4.51 & 6.15\\
LIME   
  & 0.30 & 0.41
  & 0.16 & 0.08
  & 4.16 & 5.22
  & 0.08 & 0.04
  & 1.65 & 0.58 
  & 5.22 & 6.67 \\
SHAP   
  & 0.36 & 0.40
  & 1.76 & 1.25
  & 8.57 & 5.60
  & 0.23 & 0.09
  & 1.97 & 0.46
  & 7.56 & 7.68 \\
\hline
\end{tabular}
\begin{tablenotes}[flushleft]
\footnotesize
\item \textit{Note}—The metric does not apply to Anchor because its rule-based output does not provide continuous feature importances, unlike the other two methods.
\end{tablenotes}
\end{threeparttable}
\end{table*}

\begin{table*}[!t]
\caption{Mean $\pm$ standard deviation of total (input + output) token consumption per explanation over 200 instances for three user profiles (ML engineer, domain expert, non‑technical) and three explanation methods (Anchor, LIME, SHAP) on the Heart Disease dataset (Dataset~A) and the Differentiated Thyroid Cancer Recurrence dataset (Dataset~B).}
\label{tab:tokens}
\centering
\renewcommand{\arraystretch}{1.15}
\setlength{\tabcolsep}{4pt}
\begin{tabular}{|l|*{3}{r@{\,±\,}l|}*{3}{r@{\,±\,}l|}}
\hline
\multirow{2}{*}{\textbf{Method}}
  & \multicolumn{6}{c|}{\textbf{Dataset A}}
  & \multicolumn{6}{c|}{\textbf{Dataset B}} \\
\cline{2-13}
  & \multicolumn{2}{c|}{\textbf{ML}}
  & \multicolumn{2}{c|}{\textbf{Domain}}
  & \multicolumn{2}{c|}{\textbf{Non}}
  & \multicolumn{2}{c|}{\textbf{ML}}
  & \multicolumn{2}{c|}{\textbf{Domain}}
  & \multicolumn{2}{c|}{\textbf{Non}} \\
\hline
Anchor 
  & 1131 & 133
  & 2289 & 481
  & 2314 & 480
  & 1205 & 63
  & 3358 & 287
  & 3398 & 293\\
LIME   
  & 1347 & 32
  & 2017 & 206
  & 2029 & 200
  & 1626 & 42
  & 3781 & 258 
  & 3782 & 234 \\
SHAP   
  & 1216 & 37
  & 2104 & 410
  & 2110 & 443
  & 1419 & 43
  & 3598 & 245
  & 3607 & 218 \\
\hline
\end{tabular}
\end{table*}

\begin{table*}[!t]
\centering
\caption{Satisfaction ratings (Hoffman scale: 1–5) for each explanation method (SHAP, LIME, Anchor) and user profile (ML engineer, domain expert, non‑technical) over 200 instances on the Heart Disease dataset (Dataset~A) and the Differentiated Thyroid Cancer Recurrence dataset (Dataset~B). Columns 1–7 correspond to the questionnaire items; $\bar{x}_{\text{prof}}$ is the average per profile, and $\bar{x}_{\text{meth}}$ the average per method.}
\label{tab:quality_metrics}
\renewcommand{\arraystretch}{1.12}
\setlength{\tabcolsep}{3.5pt}
\begin{tabular}{|l|l|*{9}{c|}*{9}{c|}}
\hline
\multirow{2}{*}{\textbf{Method}} 
  & \multirow{2}{*}{\textbf{Profile}}
  & \multicolumn{9}{c|}{\textbf{Dataset A}}
  & \multicolumn{9}{c|}{\textbf{Dataset B}} \\
\cline{3-20}
  & 
  & \textbf{1} & \textbf{2} & \textbf{3} & \textbf{4}
  & \textbf{5} & \textbf{6} & \textbf{7}
  & \textbf{$\bar{x}_{\text{prof}}$} & \textbf{$\bar{x}_{\text{meth}}$}
  & \textbf{1} & \textbf{2} & \textbf{3} & \textbf{4}
  & \textbf{5} & \textbf{6} & \textbf{7}
  & \textbf{$\bar{x}_{\text{prof}}$} & \textbf{$\bar{x}_{\text{meth}}$} \\
\hline
\multirow{3}{*}{SHAP}
  & ML        & 4.0 & 3.9 & 3.5 & 3.1 & 3.9 & 4.2 & 4.0 & 3.8 &  & 4.1 & 4.0 & 4.0 & 3.6 & 4.2 & 4.8 & 4.2 & 4.1 &  \\
  & Domain    & 4.2 & 4.0 & 3.7 & 3.4 & 4.0 & 4.6 & 4.4 & 4.0 & 3.9 & 4.1 & 4.0 & 3.8 & 3.3 & 4.1 & 4.3 & 4.1 & 3.9 & 4.1 \\
  & Non       & 4.1 & 4.0 & 3.3 & 3.1 & 4.0 & 4.0 & 4.0 & 3.8 &  & 4.6 & 4.3 & 3.9 & 3.6 & 4.4 & 4.1 & 4.0 & 4.1 &  \\
\hline
\multirow{3}{*}{LIME}
  & ML        & 4.0 & 3.9 & 3.5 & 3.0 & 4.0 & 4.3 & 3.9 & 3.8 &  & 4.2 & 4.1 & 3.8 & 3.6 & 4.1 & 4.6 & 4.2 & 4.0 &  \\
  & Domain    & 4.0 & 3.7 & 3.3 & 2.9 & 3.7 & 3.8 & 3.7 & 3.6 & 3.7 & 4.0 & 3.8 & 3.4 & 2.9 & 3.9 & 3.9 & 3.8 & 3.7 & 3.9 \\
  & Non       & 4.2 & 4.0 & 3.5 & 3.3 & 4.1 & 3.9 & 4.0 & 3.8 &  & 4.6 & 4.1 & 3.7 & 3.7 & 4.3 & 4.1 & 4.0 & 4.1 &  \\
\hline
\multirow{3}{*}{Anchor}
  & ML        & 3.9 & 3.5 & 3.2 & 2.8 & 3.6 & 3.7 & 3.5 & 3.5 &  & 4.0 & 3.9 & 3.7 & 3.1 & 3.9 & 4.3 & 3.9 & 3.8 &  \\
  & Domain    & 4.1 & 3.8 & 3.4 & 3.0 & 3.7 & 4.4 & 4.2 & 3.8 & 3.7 & 4.1 & 3.7 & 3.4 & 2.9 & 3.8 & 4.3 & 4.1 & 3.7 & 3.8 \\
  & Non       & 4.3 & 4.1 & 3.5 & 3.2 & 4.1 & 3.7 & 3.9 & 3.9 &  & 4.2 & 4.0 & 3.6 & 3.4 & 4.1 & 3.8 & 3.8 & 3.8 &  \\
\hline
\end{tabular}
\end{table*}

Regarding Table~\ref{tab:xai_metrics} across the three criteria---\emph{robustness} (Local Lipschitz), \emph{parsimoniousness} (Effective~Complexity) and \emph{fidelity} (Infidelity)---. LIME attains the best trade-off: the lowest Infidelity ($\approx\!0.08$--$0.30$) and the strongest robustness ($L\!<\!0.7$ Infidelity $\le 0.30$, $L<0.7$ to Heart Disease dataset), at the cost of a moderate complexity of $4$--$5$ features. Anchor produces the most parsimonious explanations: out of the 13 (or 16) available features, the model typically needs only 3–4 (Effective~Complexity) to alter its prediction. SHAP attains low‐infidelity, high‑fidelity explanations—consistent with the results reported by \cite{Qureshi2025.05.20.25327976} —yet this advantage comes at the cost of diminished robustness ($L\approx\!1.7$–$2.0$) and greater explanatory complexity ($\approx\!$ 8 features). Both drawbacks become more pronounced as the feature space expands from 13 to 16 variables.
In short, LIME offers the best balance, Anchor excels when brevity is paramount, and SHAP is preferable when capturing rich feature interactions outweighs stability considerations.

Table~\ref{tab:tokens} quantifies the total token budget (features \emph{plus} narrative) that a reader must process.  
For ML engineers Anchor is consistently the most concise \SIrange[range-phrase = {$\,\pm\,$}]{1131}{1205}{\tokens}, followed by SHAP and finally LIME, mirroring the relative verbosity of each method’s textual wrapper.  
For domain experts and non-technical users the pattern depends on the dataset: in the larger Dataset B Anchor again minimises cognitive load (\SI{7}{\%}–\SI{12}{\%} fewer tokens than SHAP, \SI{11}{\%}–\SI{14}{\%} fewer than LIME), whereas for the Dataset A, LIME required the fewest tokens overall. 
Standard deviations confirm that token counts remain stable across the 200 instances (\(\sigma \le 13\,\%\) of the mean), indicating predictable effort requirements.  
Overall, if brevity is paramount for technical stakeholders Anchor is preferable, while LIME trades additional tokens for slightly richer contextualization.

Across the seven Hoffman items (Table~\ref{tab:quality_metrics}), SHAP receives the highest mean satisfaction in both tasks (\(\bar{x}_{\text{meth}}=3.9\) on Dataset A, \(4.1\) on Dataset B). LIME trails by \(\approx0.2\) points, while Anchor ranks last yet very close (\(<0.1\) from LIME).  
Differences between user profiles are modest: Domain experts are the most critical, with an average score of 3.77, whereas non-technical users rate explanations marginally higher, especially on Dataset B. Taken together, all three XAI methods achieve solid upper-neutral acceptance (\(\ge 3.7\)), but SHAP enjoys a small, systematic advantage in perceived explanatory quality.

\section{Conclusion}
We introduced ProfileXAI, a model- and domain-agnostic framework that couples classical post-hoc explainers with retrieval-augmented LLMs to \emph{dynamically tailor} explanations to three distinct user profiles. On two medical benchmarks the system automatically chooses between SHAP, LIME and Anchor, verbalises the selected output at a suitable technical depth, and supports follow-up queries via chat.

The quantitative study confirms that no single explainer dominates every axis. LIME offers the best fidelity--robustness balance (Infidelity $\le 0.30$, $L<0.7$ to Heart Disease dataset); Anchor yields the sparsest rules and lowest token load; SHAP trades brevity for richer detail and thus achieves the highest Hoffman score ($\bar{x}=4.1$). Profile-conditioned prompts keep token use stable ($\sigma \le 13\%$) and satisfaction solidly positive ($\bar{x} \ge 3.7$), even though domain experts simulations remain the most demanding ($\bar{x}=3.77$).

These findings substantiate the value of user-adaptive narration: by aligning explanatory granularity with audience needs, ProfileXAI reconciles interpretability, cognitive economy and stakeholder satisfaction. Future work will extend the framework to multimodal data, incorporate additional explainers (e.g.\ counterfactual and concept-based), and validate with human participants to refine the simulated assessments.

\bibliographystyle{IEEEtran}
\bibliography{mybi}

\end{document}